%% file: main.tex
\documentclass[twoside,leqno,twocolumn]{article}
\usepackage{ltexpprt}
\usepackage{cite}
\usepackage{graphicx}
\usepackage{epstopdf}
\DeclareGraphicsExtensions{.eps}
\usepackage{amsmath}
\usepackage{amssymb}
\usepackage{url}
\usepackage{capt-of}
\usepackage{verbatim} 
\usepackage{soul}
\usepackage{xcolor}
\usepackage[algo2e,linesnumbered,ruled]{algorithm2e}

\newcommand{\+}[1]{\boldsymbol{#1}}
\newcommand*\samethanks[1][\value{footnote}]{\footnotemark[#1]}

\begin{document}

\title{Multi-Task Multiple Kernel Relationship Learning}
\author{Keerthiram Murugesan\thanks{School of Computer Science, Carnegie Mellon University, Pittsburgh, PA 15213, USA. $\{kmuruges,jgc\}@cs.cmu.edu$} \and Jaime Carbonell\samethanks}
\date{}
\maketitle


\begin{abstract} \small\baselineskip=9pt This paper presents a novel multitask multiple kernel learning framework that efficiently learns the kernel weights leveraging the relationship across multiple tasks. The idea is to automatically infer this task relationship in the \textit{RKHS} space corresponding to the given base kernels. The problem is formulated as a regularization-based approach called \textit{Multi-Task Multiple Kernel Relationship Learning} (\textit{MK-MTRL}), which models the task relationship matrix from the weights learned from latent feature spaces of task-specific base kernels. Unlike in previous work, the proposed formulation allows one to incorporate prior knowledge for simultaneously learning several related task. We propose an alternating minimization algorithm to learn the model parameters, kernel weights and task relationship matrix. In order to tackle large-scale problems, we further propose a two-stage \textit{MK-MTRL} online learning algorithm and show that it significantly reduces the computational time, and also achieves performance comparable to that of the joint learning framework. Experimental results on benchmark datasets show that the proposed formulations outperform several state-of-the-art multitask learning methods.
\end{abstract}

\input{Introduction}

\input{Algorithms}

\input{Experiments}

\section{Conclusion}
We proposed a novel multiple kernel multitask learning algorithm that uses inter-task relationships to efficiently learn the kernel weights. The key idea is based on the assumption that the related tasks will have similar weights for the task-specific base kernels. We proposed an iterative algorithm to jointly learn this task relationship matrix, kernel weights and the task model parameters. For large-scale datasets, we introduced a novel two-stage online learning algorithm to learn kernel weights efficiently. The effectiveness of our algorithm is empirically verified over several benchmark datasets. The results showed that both multiple kernel learning and task relationship learning for multitask problems significantly helps in boosting the performance of the model.  

\section*{Acknowledgements}
We  thank  the  anonymous  reviewers  for  their  helpful  comments.

\bibliographystyle{plain}
\bibliography{references}

\end{document}

%% file: Introduction.tex

\section{Introduction}


There have been two main lines of work in multi-task learning: First, learn a shared feature representation across all the tasks, leveraging low-dimensional subspaces in the feature space \cite{argyriou2008convex,jalali2010dirty,liu2009multi,swirszcz2012multi}. Second, learn the relationship between the tasks to improve the performance of the related tasks \cite{zhang2014regularization,zhang2010learning,rothman2010sparse,xue2007multi}. Pairwise task relationships provide very useful information for characterizing and transferring information to similar tasks.

Despite the expressive power of these two different research directions, the learning space is restricted to a single kernel (per task), chosen by the user, that corresponds to a $RKHS$ space. Multiple Kernel Learning (\textit{MKL}), on the other hand, allows the user to specify a family of base kernels related to an application, and to use the training data to automatically learn the optimal combination of these kernels. We learn the weights of the base kernels along with the model parameters in a single joint optimization problem. There is a large body of work in the recent years addressing several aspects of this problem, such as efficient learning of the kernel weights, fast optimization and providing better theoretical guarantees \cite{sun2010multiple,kloft2009efficient,kloft2011lp,lanckriet2004learning,cortes2010generalization,bach2004multiple,rakotomamonjy2008simplemkl}.

Recent work in multiple kernel learning in a multitask framework focuses on sharing common representations and assumes that the tasks are all related \cite{jawanpuria2011multi}. The motivation for this approach stems from multitask feature learning that learns joint feature representation shared across multiple tasks \cite{argyriou2008convex,swirszcz2012multi}. Unfortunately, the assumption that all the tasks are related and share a common feature representation is too restrictive for many real-world applications. Similarly, based on previous work \cite{zhang2014regularization}, one can extend the traditional multitask relationship learning \textit{MTRL} with multiple task-specific base kernels. There are two main problems with such naive approach: First, the unknown variables (task model parameters, kernel weights and task relationship matrix) are intertwined in the optimization problem, and thus making it difficult to learn for large-scale applications. Furthermore, the task relationship matrix is learned in the original feature space rather than in the kernel spaces. We show in this paper, that learning the relationship between the kernel spaces empirically performs better than relations among the original feature spaces.


There have been a few attempts to imposing higher-order relationship between kernel spaces using kernel weights. Kloft et al. \cite{kloft2011lp} propose a non-isotropic norm such as $\sqrt{\+\beta^\top \Sigma^{-1}\+\beta}$ on kernels weights $\+\beta$  to induce the relationship between the base kernels in Reproducing Kernel Hilbert Spaces. For example, in neuroimaging, a set of base kernels are derived from several medical imaging modalities such as MRI, PET etc., or image processing methods such as morphometric or anatomical modeling, etc. Since some of the kernel functions share similar parameters such as patient information, disease progression stage, etc., we can expect that these base kernels are correlated based on how they were constructed. Such information can be obtained from medical domain experts as a part of the disease prognosis which then can be used as a prior knowledge $\Sigma$. Previous work either assumes $\Sigma$ as a diagonal matrix or requires prior knowledge from the experts on the interaction of kernels \cite{kloft2011lp,hinrichs2012q}. Unfortunately, such prior knowledge is not easily available in many applications either because it is time-consuming or it is expensive to elicit.
\cite{kshirsagar2013multitask}. In such applications, we want to induce this relationship matrix from the data along with the kernel weights and model parameters.

This paper addresses these problems with a novel regularization-based approach for multitask multiple kernel learning framework, called \textit{multitask multiple kernel relationship learning} (\textit{MK-MTRL}), which models the task relationship matrix from the weights learned from the latent feature spaces of task-specific base kernels. The idea is to automatically infer task relationships in (\textit{RKHS}) spaces from their base kernels. We first propose an alternating minimization algorithm to learn the model parameters, kernel weights and task relationship matrix. The method uses a \textit{wrapper} approach which efficiently uses any off-the-shelf \textit{SVM} solver (or any kernel machine) to learn the task model parameters. However, like previous work, the proposed iterative algorithm suffers from scalability challenges. The run-time complexity of the algorithm increases with the number of tasks and the number of base kernels per task, as it needs these base kernels in memory to learn the kernel weights and the task relationship matrix. 

For large-scale applications such as object detection, we introduce a novel two-stage online learning algorithm based on recent work \cite{kumar2012binary} that learns the kernel weights independently from the model parameters. The first stage learns a good combination of base kernels in an online setting and the second stage uses the learned weights to estimate a linear combination of the base kernels, which can be readily used with a standard kernel method such as \textit{SVM} or kernel ridge regression \cite{cristianini2002kernel,cortes2010two}. We provide strong empirical evidence that learning the task relationship matrix in the RKHS space is beneficial for many applications such as stock market prediction, visual object categorization, etc. On all these applications, our proposed approach outperforms several state-of-the-art multitask learning baselines. It is worth noting that the proposed multitask multiple kernel relationship learning can be readily applied for heterogeneous and multi-view data with no modification to the proposed framework \cite{he2011graph,zhang2012inductive}.  

The rest of the paper is organized as follows: we provide a brief overview of multitask multiple kernel learning in the next section. In section \ref{mkmtrl}, we discuss the proposed model \textit{MK-MTRL}, followed by our two-stage online learning approach in section \ref{tsmkmtrl}. We then show comprehensive evaluations of the proposed model against the six baselines on several benchmark datasets in section \ref{exp}. 

\section{Preliminaries}
Before introducing our approach, we briefly review the multitask multiple kernel learning framework in this section.
Suppose there are $T$ learning tasks available with the training set $\mathcal{D}=\{(\mathbf{x}_{ti},\+y_{ti}),i=1 \ldots N_t, t=1 \ldots T \}$ , where $\+x_{ti}$ is the $i^{th}$ samples from the task $t$ and it's corresponding output $\+y_{ti}$.
Let $\{\mathbf{\mathcal{K}}_{tk}\}^{1 \leq k \leq K}$ be a set of task-specific base kernels, induced by the kernel mapping function $\phi_{k}(\cdot)$ on $t^{th}$ task data. The objective of multitask multiple kernel learning problem is to learn a \textit{good} linear combination of the task-specific base kernels $\sum_k \beta_{tk}\mathbf{\mathcal{K}}_{tk}, \beta_{tk} \geq 0$ using the relationship between the tasks.

In addition to the non-negative constraints on $\beta_{tk}$, we need to impose an additional constraint or penalty to ensure that the units in which the margins are measured are meaningful (assuming that the base kernels are properly normalized). Recent work in \textit{MKL} employs $\Vert \+\beta \Vert_2^2$ to constrain the kernel weights.  A direct extension of $\ell_2$ regularized \textit{MKL} to multi-task framework is given as follows \footnote{For clarity, we use binary classification tasks to explain the preliminaries and the proposed approach. They can be easily applied to multiclass tasks and also to regression tasks via kernel ridge regression.}: 

\begin{equation}
\begin{aligned}
\min_{\mathbf{B} \geq 0}  \min_{\mathbf{W},\mathbf{c}, \mathbf{\xi}} & \sum_{t=1}^T \Bigg(\frac{1}{2} \sum_{k=1}^K \frac{\Vert\mathbf{w}_{tk}\Vert_{\mathcal{H}_k}^2}{\beta_{tk}} + C \sum_{i=1}^{N_t} \xi_{ti} + \frac{\mu}{2}\Vert\+\beta_t\Vert_2^2\Bigg) \\
\text{s.t., }   y_{ti}(\sum_{k=1}^K & w_{tk}^\top \phi_k(\mathbf{x}_{ti})+c_t) \geq 1 - \xi_{ti}, ~ \xi_{ti} \geq 0
\label{eq:mtmkl}
\end{aligned}
\end{equation}
where ${\mathcal{H}_k}$ is the reproducing kernel Hilbert space associated with the $k^{th}$  kernel function and $\Vert \cdot \Vert_{\mathcal{H}_k}^2$ is the squared \textit{RKHS} norm.

Similarly, we can use a general $\ell_p$ norm constraint with $p>1$ on the kernel weights $(\Vert \+\beta \Vert_p^2)$. This can be thought of as a simple extension of $\ell_p$-\textit{MKL} to multi-task setting \cite{kloft2009efficient}.   Without any additional structural constraints on $\beta_{tk}$, the kernel weights are learned independently for each task and thus does not efficiently use the relationship between the tasks. Hence, we call the model in equation (\ref{eq:mtmkl}) as Independent Multiple Kernel Learning (\textit{IMKL}).

Jawanpuria and Nath \cite{jawanpuria2011multi} proposed Multi-task Multiple Kernel Feature Learning (\textit{MK-MTFL}), that employs mixed $(\ell_1,\ell_p), p\geq 2$ norm regularizer over the \textit{RKHS} norm of the feature loadings corresponding to the tasks and the base kernels. The mixed norm regularization promotes a shared feature representations to combine the given set of task-specific base kernels.  The $\ell_p$-norm regularizer learns the unequal weighting across the tasks, where as, $\ell_1$-norm regularizer over the $\ell_p$-norm leads to learning the shared kernel among the tasks.

The objective function for \textit{MK-MTFL} is given as follows:
\begin{equation}
\begin{aligned}
\min_{\mathbf{W},\mathbf{c}, \mathbf{\xi}} & \Bigg(\frac{1}{2} \sum_{k=1}^K \big(\sum_{t=1}^T\Vert\mathbf{w}_{tk}\Vert_2^p \big)^{\frac{1}{p}} \Bigg)^2 + C \sum_{t=1}^T\sum_{i=1}^{N_t} \xi_{ti} \\
\text{s.t., }  &  y_{ti}(\sum_{k=1}^K w_{tk}^\top \phi_k(\mathbf{x}_{ti})+c_t) \geq 1 - \xi_{ti}, \xi_{ti} \geq 0
\end{aligned}
\end{equation}
Note that the above objective function employs an $\ell_1$-norm across the base kernels and $\ell_p$ norm across tasks.  The above optimization problem can be equivalently written in the dual space as follows:
\begin{equation}
\begin{aligned}
\min_{\gamma \in \Delta_K} & \max_{\lambda_j \in \Delta_{T,\tilde{p}}} \max_{0 \leq \alpha \leq C} g(\lambda,\alpha,\gamma)\\
&\text{s.t., }   \mathbf{\alpha}_t^\top \mathbf{y}_t = 0,
\end{aligned}
\label{eq:mkmtfl}
\end{equation}
where,
\begin{equation*}
g(\lambda,\alpha,\gamma)=\sum_{t=1}^T \Big\{ \mathbf{1}^\top \alpha_t - \frac{1}{2}\alpha_t^\top \mathbf{Y}_t \big[ \sum_{k=1}^K \frac{\gamma_k \mathbf{\mathcal{K}}_{tk}}{\lambda_{tk}}\big] \mathbf{Y}_t \alpha_t\Big\}
\end{equation*}
Here $\+\alpha_t$ is a vector of Langragian multipliers for the $t^{th}$ task, and corresponds to $N_t$ constraints on the task data.  $\+Y_t$ is a diagonal matrix with entries as $\+y_t$ and $\mathbf{\mathcal{K}}_{tk}$ is the gram matrix of the $t^{th}$ task data w.r.t the $k^{th}$ kernel function. More specifically, $\+\gamma$ selects the base kernels that are important for all the tasks, where as $\+\lambda$ selects the base kernels that are specific to individual tasks.
With this representation, \textit{MK-MTFL} can be seen as a multiple kernel generalization to the multi-level multi-task learning proposed by Lozano and Swirszcz (2012) \cite{swirszcz2012multi}.

%% file: Algorithms.tex

\section{Multi-task Multiple Kernel Relationship Learning (\textit{MK-MTRL})}

\label{mkmtrl}
This section presents the details of the proposed model \textit{MK-MTRL}. Since multitask learning seeks to improve performance of each task with the help of other \textit{related} tasks, it is desirable in multiple kernel learning for the multitask framework to have a structural constraints on the task kernel weights $\beta_{tk}$ to promote sharing of information from other related tasks.  Note that the proposed approach is significantly different from the traditional \textit{MTRL}, as explained in the introduction. 

When prior knowledge on task relationship is available, the multiple kernel multitask learning model should incorporate this information for simultaneously learning several related tasks. Neither the \textit{IMKL} or \textit{MK-MTFL} consider the pairwise task relationship such as positive task correlation, negative task correlation, and task independence when learning the kernel weights for combining the base kernels. Based on the assumption that similar tasks are likely to give similar importance to their base kernels (and thereby, their respective \textit{RKHS} spaces), we consider a regularization on the task kernel weights  $tr(\+B \+\Omega^{-1} \+B^\top)$,  where, for notational convenience, we write $\+B= \{\+\beta_1,\+\beta_2, \ldots, \+\beta_T\}$. Mathematically, the proposed \textit{MK-MTRL} formulation is written as follows:
\begin{equation}
\begin{aligned}
\min_{\mathbf{\Omega},\mathbf{B} \geq 0}  \min_{\mathbf{W},\mathbf{c}, \mathbf{\xi}} & \sum_{t=1}^T \Bigg(\frac{1}{2} \sum_{k=1}^K \frac{\Vert\mathbf{w}_{tk}\Vert_{\mathcal{H}_k}^2}{\beta_{tk}} + C \sum_{i=1}^{N_t} \xi_{ti}\Bigg) \\
& + \frac{\mu}{2}tr(\mathbf{B}\mathbf{\Omega}^{-1}\mathbf{B}^\top)\\
\text{s.t., }   y_{ti}(\sum_{k=1}^K & w_{tk}^\top \phi_k(\mathbf{x}_{ti})+c_t) \geq 1 - \xi_{ti}, \xi_{ti} \geq 0\\
& \mathbf{\Omega} \succeq 0,\\
& tr(\mathbf{\Omega}) \leq 1
\label{eq:mkmtrl}
\end{aligned}
\end{equation}
The key difference from the \textit{IMKL} model is that the standard (squared) $\ell_p$ norm on $\+\beta_t$ is replaced with a more meaningful structural penalty that incorporates the task relationship. Unlike in \textit{MK-MTFL}, the shared information among the task is separate from the core problem ($T$ SVMs).  Here, $\+\Omega$ encodes the task relationship such that similar tasks are forced to have similar kernel weights.
It is easy to see that when $\+\Omega=\mathbb{I}_{T\times T}$, the above problem reduces to equation (\ref{eq:mtmkl}).

\subsection{MK-MTRL in Dual Space}
In this section, we consider the proposed approach in the dual space. By writing the above objective function in Lagrangian form and introducing Lagrangian multiplier $\alpha_{tk}$ for the constraints, we can write the corresponding dual objective function as:   
\begin{equation}
\begin{aligned}
\min_{\mathbf{\Omega},\mathbf{B} \geq 0} \max_{0 \leq \alpha \leq C} & h(\alpha,\mathbf{B}) + \frac{\mu}{2}tr(\mathbf{B}\mathbf{\Omega}^{-1}\mathbf{B}^\top)\\
\text{s.t., }  & \mathbf{\alpha}_t^\top \mathbf{y}_t = 0,\\
& \mathbf{\Omega} \succeq 0,\\
& tr(\mathbf{\Omega}) \leq 1
\label{eq:mkmtrldual}
\end{aligned}
\end{equation}
where,
\begin{equation*}
 h(\alpha,\mathbf{B})=\sum_{t=1}^T  \Big\{\mathbf{1}^\top \mathbf{\alpha}_t -\frac{1}{2} \mathbf{\alpha}_t^\top \mathbf{Y}_t \Big(\sum_{k=1}^K \beta_{tk}\mathbf{\mathcal{K}}_{tk}\Big) \mathbf{Y}_t \mathbf{\alpha}_t \Big\}
\end{equation*}
The above objective function is a bi-convex optimization problem.  Note that we can further reduce the problem by eliminating $\alpha_{tk}$, then the dual problem becomes:

\begin{equation}
\begin{aligned}
\min_{\mathbf{\Omega}} \max_{0 \leq \alpha \leq C} \sum_{t=1}^T & \Big\{\mathbf{1}^\top \mathbf{\alpha}_t -\frac{1}{2} \Vert \mathbf{\mathcal{G}} \Vert_{\Omega} \Big\} \\
\text{s.t., }  & \mathbf{\alpha}_t^\top \mathbf{y}_t = 0,\\
& \mathbf{\Omega} \succeq 0,\\
& tr(\mathbf{\Omega}) \leq 1
\end{aligned}
\end{equation}
where $\mathbf{\mathcal{G}}_{tk} =\beta_{tk} \mathbf{\alpha}_t^\top \mathbf{Y}_t \mathbf{\mathcal{K}}_{tk} \mathbf{Y}_t \mathbf{\alpha}_t$ which corresponds to $\frac{\Vert\+w_{tk}\Vert_2^2}{\beta_{tk}}$ in the primal space and we write $\Vert \mathbf{\mathcal{G}} \Vert_{\Omega} = \sqrt{tr(\mathbf{\mathcal{G}} \+\Omega\mathbf{\mathcal{G}}^{\top})}$. We will use this representation for deriving closed-form solution for the task kernel weights $\+B$

\subsection{Optimization}
We use an alternating minimization procedure for learning the kernel weights and the model parameters iteratively. We implement a two-layer \textit{wrapper} approach commonly used in these \textit{MKL} solvers for our problem. The wrapper methods alternate between minimizing the primal problem (\ref{eq:mkmtrl}) w.r.t $\+\beta_t$ via a simple analytical update step and minimizing all other variables in terms of the dual variables $\+\alpha_t$ from equation (\ref{eq:mkmtrldual}).

When $\{\+B, \+\Omega\}$ are fixed, \textit{MK-MTRL} equation (\ref{eq:mkmtrldual})  reduces to $T$ independent sub-problems. One can use any conventional \textit{SVM} solver (or any kernel method) to optimize for $\+\alpha_t$ independently. We focus on optimizing the kernel coefficients $\+B$ and $\+\Omega$ next. 

\subsubsection*{Optimizing w.r.t $\+B$ when $\{\+\alpha, \+\Omega\}$ are fixed }
Given $\{\+\alpha, \+\Omega\}$, we find $\+B$ by setting the gradient of equation (\ref{eq:mkmtrl}) w.r.t $\+B$ to zero and we get:
\begin{equation}
 \begin{aligned} \label{eq:muom}
      \+B=\frac{1}{\mu}(\mathcal{W}\circ\+B^{-2})\+\Omega 
      \end{aligned}
\end{equation}
where $\+B^{-2}=\{\beta^{-2}_{kt}, 1 \leq k \leq K, 1 \leq t \leq T\}$, $\mathcal{W}_{tk}=\Vert\mathbf{w}_{tk}\Vert_{\mathcal{H}_k}^2$ and $\+A \circ \+B$ is an element-wise product operation.

By incorporating the last term in equation (\ref{eq:mkmtrl}) into the constraint set, we can eliminate the regularization parameter $\mu$ to obtain an analytical solution for $\+B$. Because $\+\Omega \succeq 0$ and $\+B \geq 0$, the constraint $tr(\+B\+\Omega^{-1}\+B^\top) \leq 1$ must be active at optimality. We can now use the above equation to solve for $\mu$. 
\begin{equation}
 \begin{aligned} \label{eq:omcf}
      \+B=\frac{(\mathcal{W}\circ\+B^{-2})\+\Omega}{\sqrt{tr((\mathbf{\mathcal{W}}\circ\+B^{-2})\+\Omega (\mathcal{W}\circ\+B^{-2})^\top)}}
      \end{aligned}
\end{equation}

Since the task relationship matrix is independent of the number of base kernels $K$, one may use the above closed-form solution when the number of tasks is small. For some applications, it may be desirable to employ an iterative approach such as first-order method (\textit{FISTA}) or second-order method (\textit{Newton's}). The parameter $\mu$ can be easily learned by cross-validation.
\subsubsection*{Optimizing w.r.t $\+\Omega$ when $\{\+\alpha, \+B\}$ are fixed }
In the final step of the optimization, we fix $\+\alpha$ and $\+B$ and solve the problem w.r.t $\+\Omega$. By taking the partial derivative of the objective function with respect to $\+\Omega$ and setting it zero, we get an analytical solution for $\+\Omega$\cite{zhang2014regularization}:
\begin{equation}
 \begin{aligned} \label{eq:bcf}
      \+\Omega=\frac{(\+B^{\top}\+B)^{\frac{1}{2}}}{tr((\+B^{\top}\+B)^{\frac{1}{2}})}
      \end{aligned}
\end{equation}
Substituting the above solution in equation \ref{eq:mkmtrl}, we can see that the the objective function of \textit{MK-MTRL} is related to the trace norm regularization.  Instead of $\ell_p$ norm regularization (as in $L_p$-MKL) or mixed-norm regularization (as in \textit{MK-MTFL}), our model seeks a low-rank $\+B$, using $\Vert \+B\Vert_*$, such that similar base kernels are selected among the similar tasks.

\section{Two-Stage Multi-task Multiple Kernel Relationship Learning}
\label{tsmkmtrl}
The proposed optimization procedure in the previous section involves $T$ independent $SVM$ (or \textit{kernel ridge regression}) calls, followed by two closed-form expressions for jointly learning the kernel weights $\+B$, task relationship matrix $\+\Sigma$ and the task parameters $\+\alpha$. Even though this approach is simple and easy to implement, it requires the precomputed kernel matrices to be loaded into memory for learning the kernel weights. This could add a serious computational burden especially when the number of tasks $T$ is large \cite{weinberger2009feature}.

In this section, we consider an alternative approach to address this problem inspired by \cite{cristianini2002kernel,cortes2010two}. It follows a two-stage approach: first, we independently learn the weights of the given task-specific base kernels using the training data and then, we use the weighted sum of these base kernels in a standard kernel machines such as $SVM$ or \textit{kernel ridge regression} to obtain a classifier. This approach significantly reduces the amount of computational overhead involved in the traditional multiple kernel learning algorithms that estimate the kernel weights and the classifier by solving a joint optimization problem. 

We propose an efficient binary classification framework for learning the weights of these task-specific base kernels, based on target alignment \cite{cristianini2002kernel}. The proposed framework formulates the kernel learning problem as a linear classification in the kernel space (so called $\mathcal{K}$-classifier). In this space, any task classifier with weight parameters directly corresponds to the task kernel weights.

	\begin{algorithm2e}[ht]
	
	\SetKwInOut{Input}{Input}
	\SetKwInOut{Output}{Output}
	\Input{$Base~kernels ~\{\mathbf{\mathcal{K}}_{tk}\}_{1 \leq t \leq T}^{1 \leq k \leq K}$, $labels ~ \{\+y_t)\}_{t=1}^T$, $regularization~~ parameter~ \mu > 0$}
	\Output{$\+\alpha,\+B, \+\Omega$}
	\textit{Initialize} $\+\Omega=\frac{1}{T}\mathbb{I}_{T\times T}$\;
	\Repeat{converges}
	{
	\Repeat{converges}
	{
	\textit{Set} $\mathbf{\mathcal{K}}_t \gets \sum_{k=1}^K \beta_{tk}\mathbf{\mathcal{K}}_{tk}, \forall t \in [T]$;
	
	\textit{Solve} for $\+\alpha_t, t \in [T]$ 
	\begin{equation}
	\max_{0 \leq \+\alpha_t\leq C, \+\alpha_t^\top\+y_t=0} \Big\{\mathbf{1}^\top \mathbf{\alpha}_t -\frac{1}{2} \mathbf{\alpha}_t^\top \mathbf{Y}_t \mathcal{K}_t \mathbf{Y}_t \mathbf{\alpha}_t \Big\} ~~ \text{(\textit{SVM})}
	\label{eq:libsvm}
	\end{equation}
	
	\textit{Solve} for $\+B=\{\+\beta_1,\+\beta_2, \ldots, \+\beta_T\},$ 
	\begin{equation}
	\min_{\mathbf{B} \geq 0}  \frac{1}{2} \sum_{t=1}^T \sum_{k=1}^K \frac{\Vert\mathbf{w}_{tk}\Vert_{\mathcal{H}_k}^2}{\beta_{tk}}  + \frac{\mu}{2} tr(\mathbf{B}\mathbf{\Omega}^{-1}\mathbf{B}^\top)
	\end{equation}
	$~~~$ where $\Vert\mathbf{w}_{tk}\Vert_{\mathcal{H}_k}^2 = \beta_{tk}^2\mathbf{\alpha}_t^\top \mathbf{Y}_t \mathcal{K}_{tk} \mathbf{Y}_t \mathbf{\alpha}_t$
	}
	\textit{Solve} for $\+\Omega,$
	\begin{equation}
	\min_{\+\Omega \succeq \+0, tr(\+\Omega) \leq 1} tr(\+\Omega^{-1}(\+B^\top\+B))
	\label{eq:svd}
	\end{equation}
	
	}
	\caption{Wrapper method for Multi-task Multiple Kernel Relationship Learning (\textit{MK-MTRL})}
	\label{alg:mkmtrl}
	\end{algorithm2e}

		\begin{algorithm2e}[ht]
			
			\SetKwInOut{Input}{Input}
			\SetKwInOut{Output}{Output}
			\Input{$Base~kernels ~\{\mathbf{\mathcal{K}}_{tk}\}_{1 \leq t \leq T}^{1 \leq k \leq K}$, $labels ~ \{\+y_t)\}_{t=1}^T$, $regularization~~ parameter~ \mu > 0$, Number of rounds $R$}
			\Output{$\+\alpha,\+B, \+\Omega$}
			\textbf{Initialize} $\+\beta_t^{(1)}=\+0, \+\Omega=\frac{1}{T}\mathbb{I}_{T\times T}$\;
			
			\For{$r = 1 \ldots R$}{
				
			\textit{Construct} $(z_{t,ii'},l_{t,ii'})$ using $\mathcal{K}$ for any two examples $(\mathbf{x}_{ti},y_{ti})$ and $(\mathbf{x}_{ti'},y_{ti'})$ and for any task $t$, where
			\begin{equation}
			\begin{aligned}
			z_{t,ii'} &=\{\mathcal{K}_1(\mathbf{x}_{ti},\mathbf{x}_{ti'}),\mathcal{K}_2(\mathbf{x}_{ti},\mathbf{x}_{ti'}), \\
			&\ldots, \mathcal{K}_K(\mathbf{x}_{ti},\mathbf{x}_{ti'})\}\\
			l_{t,ii'} &= 2. \mathbf{1}\{y_{ti}=y_{ti'}\}-1
			\end{aligned}
			\end{equation}
			
			\textit{Predict} $\hat{l}_{t,ii'}=\+\beta_t^{(r)\top} z_{t,ii'}$
			
			\If{$(l_{t,ii'} \neq \hat{l}_{t,ii'})$}
			{
				\For{$t' = 1 \ldots T$}{
					$\beta_{t'}^{(r+1)}=\beta_{t'}^{(r)} + \frac{1}{\mu}l_{t,ii'}\+\Omega_{t,t'} z_{t,ii'}$
				}
				\textit{Solve} for $\+\Omega,$
				\begin{equation}
				\min_{\+\Omega \succeq \+0, tr(\+\Omega) \leq 1} tr(\+\Omega^{-1}(\+B^\top\+B))
				\end{equation}
			}
		}
			\textit{Set} $\mathbf{\mathcal{K}}_t \gets \sum_{k=1}^K \beta_{tk}^{(R)}\mathbf{\mathcal{K}}_{tk}, \forall t \in [T]$;
			
			\textit{Solve} for $\+\alpha_t, t \in [T]$ 
			\begin{equation}
				\max_{0 \leq \+\alpha_t\leq C, \+\alpha_t^\top\+y_t=0} \Big\{\mathbf{1}^\top \mathbf{\alpha}_t -\frac{1}{2} \mathbf{\alpha}_t^\top \mathbf{Y}_t \mathcal{K}_t \mathbf{Y}_t \mathbf{\alpha}_t \Big\} ~~ \text{(\textit{SVM})}
			\end{equation}

			\caption{Two-stage, online learning of (\textit{MK-MTRL})}
			\label{alg:2gmkmtrl}
		\end{algorithm2e}
		
For a given set of $T*K$ base kernels $\{\mathbf{\mathcal{K}}_{tk}\}_{1 \leq t \leq T}^{1 \leq k \leq K}$ ($K$ base kernels per task), we define a binary classification framework over a new instance space (so called $\mathcal{K}$-space) defined as follows:
\begin{equation}
\begin{aligned}
z_{t,ii'} &=\{\mathcal{K}_1(\mathbf{x}_{ti},\mathbf{x}_{ti'}),\mathcal{K}_2(\mathbf{x}_{ti},\mathbf{x}_{ti'}), \ldots, \mathcal{K}_K(\mathbf{x}_{ti},\mathbf{x}_{ti'})\}\\
l_{t,ii'} &= 2. \mathbf{1}\{y_{ti}=y_{ti'}\}-1
\end{aligned}
\end{equation}

Any hypothesis $h_t:\mathbb{R}^K \rightarrow \mathbb{R}$ for a task $t$ induces a similarity function $
\tilde{\mathcal{K}}_{h_t}(\mathbf{x}_{ti},\mathbf{x}_{ti'})$ between instances $\mathbf{x}_{ti}$ and $\mathbf{x}_{ti'}$ in the original space:

\begin{equation*}
\begin{aligned}
 &\tilde{\mathcal{K}}_{h_t}(\mathbf{x}_{ti},\mathbf{x}_{ti'}) = h_t(z_{t,ii'}) \\
 &= h_t(\mathcal{K}_1(\mathbf{x}_{ti},\mathbf{x}_{ti'}), \ldots, \mathcal{K}_K(\mathbf{x}_{ti},\mathbf{x}_{ti'}))
 \end{aligned}
\end{equation*}

Suppose we consider a linear function for our task hypothesis $h_t(z_{t,ii'})=\+\beta_t.z_{t,ii'}$ with the non-negative constraints $\+\beta_t \geq 0$, then the resulting induced kernel $ \tilde{\mathcal{K}}_{h_t}$ is also positive semi-definite.  The key idea behind this two-stage approach is that if a $\mathcal{K}$-classifiers $h_t$ is a good classifier in the $\mathcal{K}$-space, then the induced kernel $\tilde{\mathcal{K}}_{h_t}(\mathbf{x}_{ti},\mathbf{x}_{ti'})$ will likely be positive when $\mathbf{x}_{ti}$ and $\mathbf{x}_{ti'}$ belong to the same class and negative otherwise. Thus the problem of learning a good combination of base kernels can be framed as a problem of learning a good $\mathcal{K}$-classifier. 

With this framework, the optimization problem for learning $\+\beta_t$ for each task $t$ can be formulated as follows:
\begin{equation}
\begin{aligned}
& \min_{\mathbf{B} \geq 0} \sum_{t=1}^T  \ell(l_{t,ii'},\langle \beta_t, z_{t,ii'} \rangle) +  \frac{\mu}{2} \mathcal{R}(\mathbf{B})\\
& \ell(l_{t,ii'},\langle \beta_t, z_{t,ii'} \rangle) = \frac{1}{\binom{N_t}{2}+N_t}\sum_{1 \leq i \leq i' \leq N_t} \big[ 1-l_{t,ii'}\mathbf{\beta}_t z_{t,ii'}\big]_+
\end{aligned}
\end{equation}
where $[1-s]_+ = \max\{0,1-s\}$ and $\mathcal{R}(\mathbf{B})$ is the regularization function on the kernel weights $\+B$. Since we are interested in learning task relationships using the task kernel weights $\+\beta_t$, we can directly extend the above formulation to incorporate the regularization on $\+\beta_t$ based on \textit{MK-MTRL}.
\begin{equation}
\begin{aligned}
\min_{\mathbf{\Omega}} \min_{\mathbf{B} \geq 0} \sum_{t=1}^T &  \ell(l_{t,ii'},\langle \beta_t, z_{t,ii'} \rangle) +  \frac{\mu}{2}tr(\mathbf{B}\mathbf{\Omega}^{-1}\mathbf{B}^\top)\\
	& \mathbf{\Omega} \succeq 0,\\
	& tr(\mathbf{\Omega}) \leq 1
\end{aligned}
\end{equation}
Since the above objective function depends on every pair of observations, we consider an online learning procedure for faster computation that learns the kernel weights and the task relationship matrix sequentially. Due to space limitations, we show the online version of our algorithms in the supplementary section. Note that with the above formulation, one can easily extend the existing approach to jointly learn both the feature and task relationship matrices using matrix normal penalty \cite{zhang2010learning}.

\section{Algorithms}
		
	Algorithm \ref{alg:mkmtrl} shows the pseudo-code for \textit{MK-MTRL}. It outlines the update steps explained in Section 3. The algorithm alternates between learning the model parameters, kernel weights and task relationship matrix until it reaches the maximum number of iterations \footnote{\textit{maxIter} is set to $50$} or when there are minimal changes in the subsequent $\+B$.

	The two-stage, online learning of \textit{MK-MTRL} is given in Algorithm \ref{alg:2gmkmtrl}. The online learning of $\+\beta_t$ and $\+\Omega$ is based on the recent work by Saha et. al., 2011 \cite{saha2011online}. We set the maximum number of rounds to $100,000$. Since we construct the examples in kernel space on the fly, there is no need keep the base kernel matrices in memory.  This significantly reduces the computational burden required in computing $\+B$.
	
	We use \textit{libSVM} to solve the $T$ individual SVMs (equation \ref{eq:libsvm}). All the base kernels are normalized to unit trace. Note that equation \ref{eq:svd} requires computing Singular Value Decomposition (SVD) on $(\+B^{\top}\+B)$. One may use an efficient decomposition algorithm such as the randomized $SVD$ to speed up the learning process \cite{liberty2007randomized}.

%% file: Experiments.tex

\section{Experiments}

\label{exp}

\begin{table*}[ht]
	\renewcommand{\arraystretch}{1.25}
	\centering
	\caption{Mean Squared Error (MSE) for each company $(\times 1000)$}
	\label{stockres}
	\begin{tabular}{l|llll|lllll|}
		\cline{2-10}
		& OLS  & Lasso & MRCE & FES  & STL  & IKL  & IMKL  & MK-MTFL & MK-MTRL \\ \hline
		\multicolumn{1}{|l|}{Walmart}        & 0.98 & 0.42  & 0.41 & 0.40 & 0.44 & 0.43 & 0.45 & 0.44    & 0.44          \\
		\multicolumn{1}{|l|}{Exxon}          & 0.39 & 0.31  & 0.31 & 0.29 & 0.34 & 0.32 & 0.33 & 0.32    & 0.32          \\
		\multicolumn{1}{|l|}{GM}             & 1.68 & 0.71  & 0.71 & 0.62 & 0.82 & 0.62 & 0.60 & 0.61    & \textbf{0.56} \\
		\multicolumn{1}{|l|}{Ford}           & 2.15 & 0.77  & 0.77 & 0.69 & 0.91 & 0.56 & 0.53 & 0.55    & \textbf{0.49} \\
		\multicolumn{1}{|l|}{GE}             & 0.58 & 0.45  & 0.45 & 0.41 & 0.43 & 0.41 & 0.40 & 0.40    & \textbf{0.39} \\
		\multicolumn{1}{|l|}{ConocoPhillips} & 0.98 & 0.79  & 0.79 & 0.79 & 0.84 & 0.81 & 0.83 & 0.80    & 0.80          \\
		\multicolumn{1}{|l|}{Citigroup}      & 0.65 & 0.66  & 0.62 & 0.59 & 0.64 & 0.66 & 0.62 & 0.62    & 0.60          \\
		\multicolumn{1}{|l|}{IBM}            & 0.62 & 0.49  & 0.49 & 0.51 & 0.48 & 0.47 & 0.45 & 0.45    & \textbf{0.43}          \\
		\multicolumn{1}{|l|}{AIG}            & 1.93 & 1.88  & 1.88 & 1.74 & 1.91 & 1.94 & 1.88 & 1.89    & 1.83          \\ \hline
		\multicolumn{1}{|l|}{AVG}                                  & 1.11 & 0.72  & 0.71 & 0.67 & 0.76 & 0.69 & 0.68 & 0.68    & \textbf{0.65}          \\ \hline
	\end{tabular}
\end{table*}

 We evaluate the performance of our proposed model on several benchmark datasets. We compare our proposed model with five state-of-the-art baselines in multitask learning and in multitask multiple kernel learning. All reported results in this section are averaged over $10$ random runs of the training data. Unless otherwise specified, all model parameters are chosen via 5-fold cross validation. The best model and models with statistically comparable results are shown in bold.

\subsection{Compared Models}
We compare the following models for our evaluation.
 \begin{itemize}
 	\item Single-Task Learning (STL) learns the tasks independently. STL uses either \textit{SVM} (in case of binary classification tasks) or Kernel Ridge regression (in case of regression tasks) to learn the individual models.
 	\item Multi-task Feature Learning (MTFL \cite{argyriou2008convex}) learns a shared feature representation from all the tasks using regularization. It learns this shared feature representation along with the task model parameters alternatively\footnote{The source code for this baseline is available at \url{http://ttic.uchicago.edu/~argyriou/code/mtl_feat/mtl_feat.tar}}.
 	\item Multi-task Relationship Learning (MTRL \cite{zhang2014regularization}) learns task relationship matrix under a regularization framework. This model can be viewed as a multitask generalization for single-task learning. It learns the task relationship matrix and the task parameters in an iterative fashion\footnote{The source code for this baseline is available at \url{https://www.cse.ust.hk/~zhangyu/codes/MTRL.zip}}.
 	\item Single-task Multiple Kernel Learning (IMKL) learns independent MKL for each task. This baseline does not use any shared information between the tasks. We use $\ell_p$-MKL for each task. We tune the value of $p$ from $[2,3,4,6,8.67]$ using $5$-fold cross validation.
 	\item Multi-task Multiple Kernel Feature Learning (MK-MTFL \cite{jawanpuria2011multi}) learns a shared kernel for feature representation from all tasks. This is a multiple kernel generalization of multitask feature learning problem. Again, we tune the value of $\tilde{p}$ from $[2,3,4,6,8.67]$ using $5$-fold cross validation\footnote{The source code for this baseline is available at \url{http://www.cse.iitb.ac.in/saketh/research/MTFL.tgz}}.
 \end{itemize}
 
 Unless otherwise specified, the kernels for \textit{STL}, \textit{MTFL} and \textit{MTRL} are chosen (via cross validation) from either a Gaussian RBF kernel with different bandwidth or a linear kernel for each dataset. The value for $C$ is chosen from $[10^{-3}, \ldots, 10^{3}]$. We tune the value of $\mu$ from $[10^{-7}, \ldots, 10^3]$. We use \textit{Newton's} method to learn the task kernel weight matrix $\+B$ for the alternating minimization algorithm. We compare our models on several applications: Asset Return Prediction, Landmine Detection and Object Recognition \footnote{See supplementary material for additional experiments}. It is worth noting that different applications require different types of base kernels and there is no common set of kernel functions that will work for all applications. We choose these base kernels based on the application and the type of data.

\subsection {Asset Return Prediction}  We begin our experiments with asset return prediction data used in \cite{rothman2010sparse} {\footnote{\url{http://cran.r-project.org/web/packages/MRCE/index.html}}}. It consists of weekly log returns of $9$ stocks from the year $2004$. This dataset is considered in linear multivariate regression with output covariance estimation techniques \cite{rothman2010sparse}. We consider first-order vector auto-regressive models of the form ${\+x_t}=f(\+x_{t-1})$ where $\+x_t$ corresponds to the $9$-dimensional vector of weekly log-returns from $9$ companies as shown in table \ref{stockres}.  The dataset is split evenly such that the first $26$ weeks of the year is used as the training set and the next $26$ weeks is used as the test set. Following \cite{sindhwani2012scalable}, we use univariate Gaussian kernels with $13$ varying bandwidth, generated from each feature, as base kernels. The total number of base kernels sums to $117$.

Performance is measured by the average mean-squared prediction error over the test set for each task. The experimental setup for this dataset follows exactly \cite{rothman2010sparse}. We compare the results from our proposed and baseline model with the results from Ordinary Least Square (\textit{OLS}), Lasso, Multivariate Regression with Covariate Estimation (\textit{MRCE}) and Factor Estimation and Selection (\textit{FES}) models reported in \cite{rothman2010sparse} (See \cite{rothman2010sparse} for more details about the models). In addition to the standard baselines, we include Input Kernel Learning (\textit{IKL}), which learns a vector of kernel weights $\+\beta$ shared by all tasks \cite{tang2009multiple}. 

After running \textit{MK-MTRL} on these $117$ base kernels, the model sets most of them to $0$ except for base kernels corresponding to bandwidths ($1e-4,1$). These bandwidth selections represent the long-term and short-term dependencies common in temporal data. We reran the model with these selected non-zero bandwidths and report the results for these selected base kernels. We can see that the proposed model \textit{MK-MTRL} performs better than all the baselines.

\begin{table*}[ht]
\renewcommand{\arraystretch}{1.5}
\centering
\caption{Average $AUC$ scores for different samples of \textit{landmine} dataset. The table reports the mean and standard errors over 10 random runs.}
\label{tab:landmine}
\begin{tabular}{l|l|l|l|}
\cline{2-4}
                             & 30 samples                  & 50 Samples                  & 80 Samples                  \\ \hline
\multicolumn{1}{|l|}{STL}    & 0.6315 $\pm$ 0.032          & 0.6540 $\pm$ 0.026          & 0.6542 $\pm$ 0.027          \\
\multicolumn{1}{|l|}{MTFL}   & 0.6387 $\pm$ 0.037          & 0.6968 $\pm$ 0.015          & 0.7051 $\pm$ 0.020          \\
\multicolumn{1}{|l|}{MTRL}   & 0.6555 $\pm$ 0.034          & 0.6933 $\pm$ 0.023          & 0.7074 $\pm$ 0.024          \\
\multicolumn{1}{|l|}{IMKL}    & 0.6857 $\pm$ 0.024          & 0.7138 $\pm$ 0.011          & 0.7278 $\pm$ 0.011          \\
\multicolumn{1}{|l|}{MK-MTFL} & \textbf{0.6866 $\pm$ 0.018} & 0.7145 $\pm$ 0.009          & 0.7305 $\pm$ 0.009          \\
\multicolumn{1}{|l|}{MK-MTRL} & \textbf{0.6870 $\pm$ 0.033} & \textbf{0.7242 $\pm$ 0.011} & \textbf{0.7405 $\pm$ 0.014} \\ \hline
\end{tabular}
\end{table*}

\subsection{Landmine Detection}
This dataset \footnote{\url{http://www.ee.duke.edu/~lcarin/LandmineData.zip}} consists of $19$ tasks collected from different landmine fields. Each task is a binary classification problem: landmines $(+)$ or clutter $(-)$ and each example consists of $9$ features extracted from radar images with four moment-based features, three correlation-based features, one energy ratio feature and a spatial variance feature. Landmine data is collected from two different terrains:  tasks $1-10$ are from highly foliated regions and tasks $11-19$ are from desert regions, therefore tasks naturally form two clusters. Any hypothesis learned from a task should be able to utilize the information available from other tasks belonging to the same cluster.

We choose $\{30,50,80\}$ examples per task for this dataset. We use a polynomial kernel with power $\{1,2,3,4,5\}$ for generating our base kernels. Note that we intentionally kept the size of the training data small to drive the need for learning from other tasks, which diminishes as the training sets per task become large. Due to class-imbalance issue (with few $(+)$ examples compared to $(-)$ examples), we use average {Area Under the ROC Curve} ($AUC$) as the performance measure. This dataset has been previously used for jointly learning feature correlation and task correlation \cite{zhang2010learning}. Hence, \textit{landmine} dataset is an ideal dataset for evaluating all the models. 

Table \ref{tab:landmine} reports the results from the experiment. We can see that \textit{MK-MTRL} performs better in almost all cases. When the number of training examples is small, \textit{MK-MTRL} has difficulty in learning the task relationship matrix $\+\Omega$ as it depends on the kernel weights. On the other hand, \textit{MK-MTFL} performs equally well as it shares the feature representation among the tasks which is especially useful when the number of training is relatively low. As we get more and more training data, \textit{MK-MTRL} performs significantly better than all the other baselines.

\subsection{Object Recognition}

In this section, we evaluate our two proposed algorithms for \textit{MK-MTRL} with computer vision datasets, \textit{Caltech101}\footnote{\url{http://www.vision.ee.ethz.ch/~pgehler/projects/iccv09}} and \textit{Oxford} flowers \footnote{\url{http://www.robots.ox.ac.uk/~vgg/data/flowers/17/datasplits.mat}} in terms of accuracy and training time. \textit{Caltech101} dataset consists of $9,144$ images from $102$ categories of objects such as faces, watches, animals, etc. The minimum, average and maximum number of images per category are $31$,$90$ and $800$ respectively. The \textit{Caltech101} base kernels for each task are generated from feature descriptors such as geometric blur, PHOW gray/color, self-similarity, etc. For each of the $102$ classes, we select $30$ examples (for a total of $3060$ examples per task) and then split these $30$ examples into testing and training folds, which ensures matching training and testing distributions. \textit{Oxford} flowers consists of $17$ varieties of flowers and the number of images per category is $80$. The \textit{Oxford} base kernels for each task are generated from a subset of feature values. Each one-vs-all binary classification problem is considered as a single task, which amount to $102$ and $17$ tasks with $38$ and $7$-base kernels per task, respectively. Following the previous work, we set the value of $C=1000$ for \textit{Caltech101} dataset.

\begin{figure*}[ht]
	\centering
	\includegraphics[width=3in]{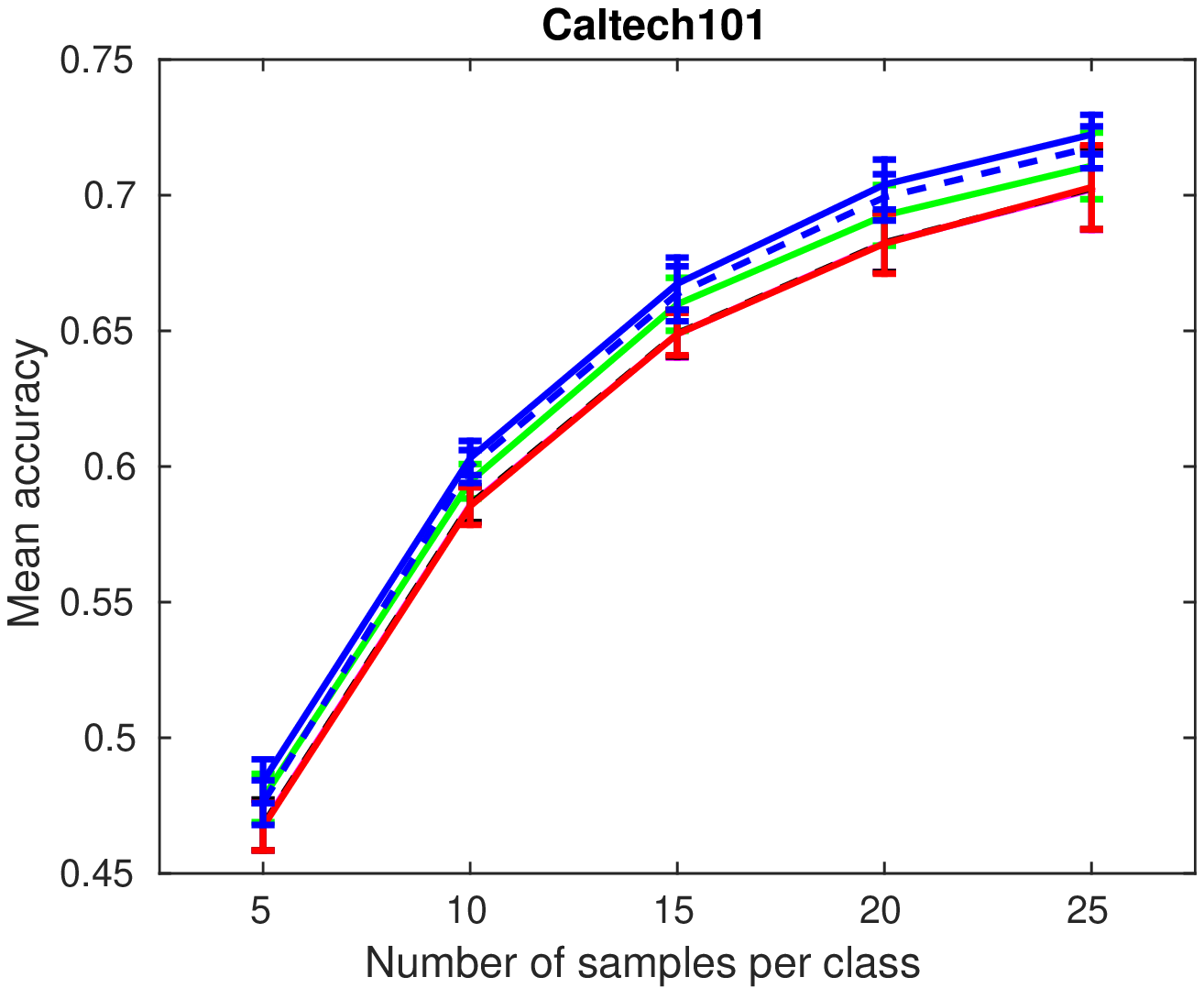}
	\hspace{0.5em}
	\includegraphics[width=3in]{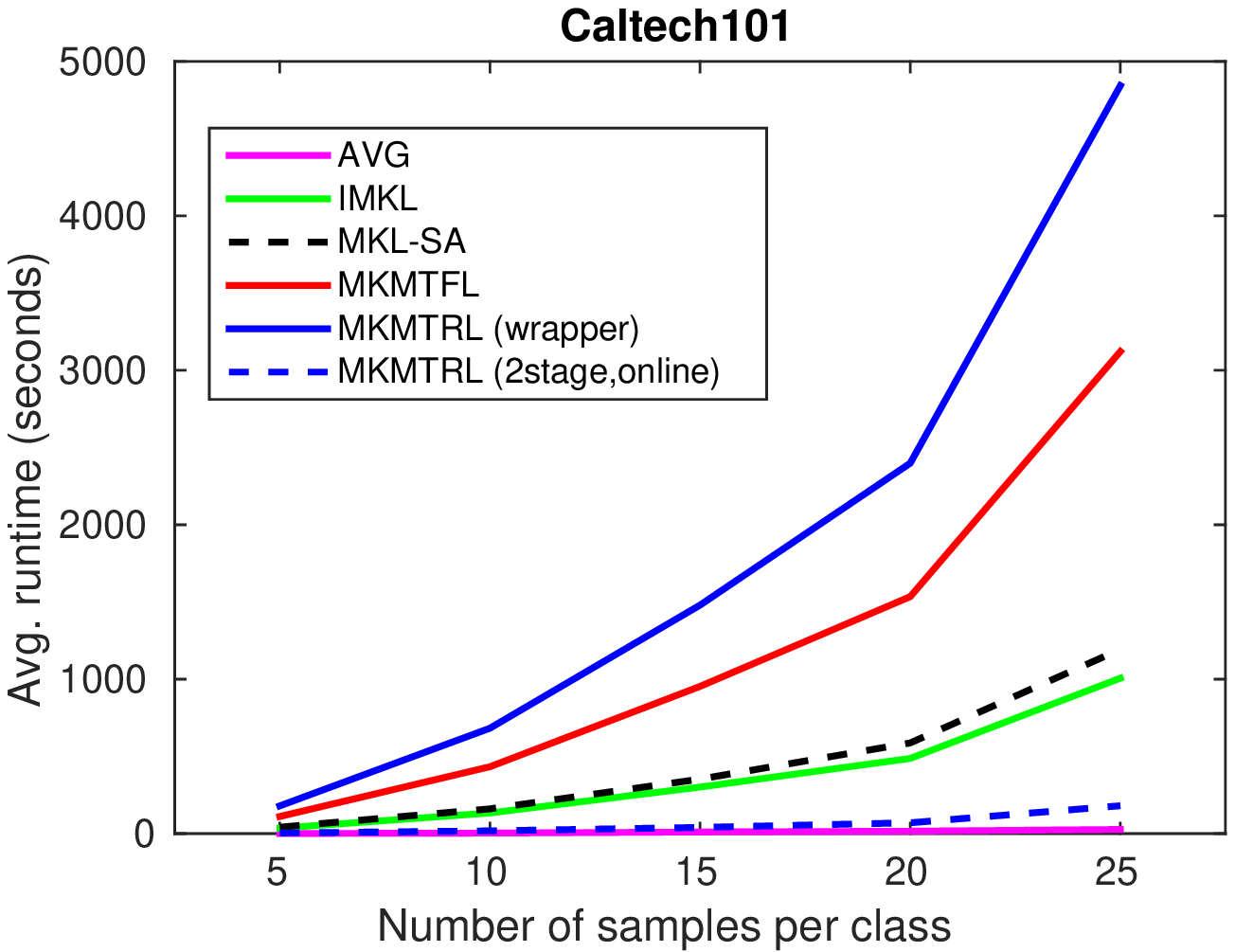}

	\includegraphics[width=3in]{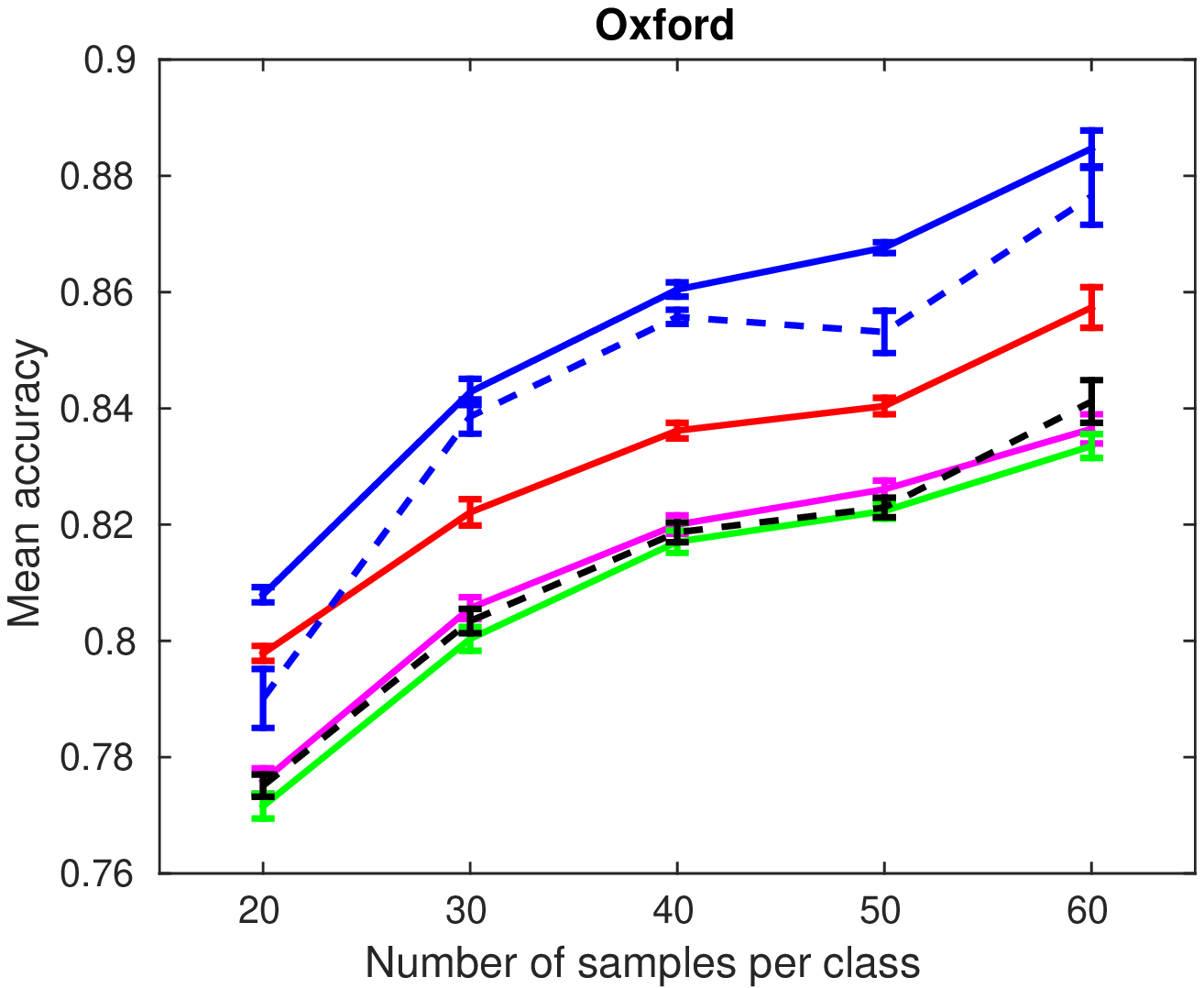}
	\hspace{0.5em}
	\includegraphics[width=3in]{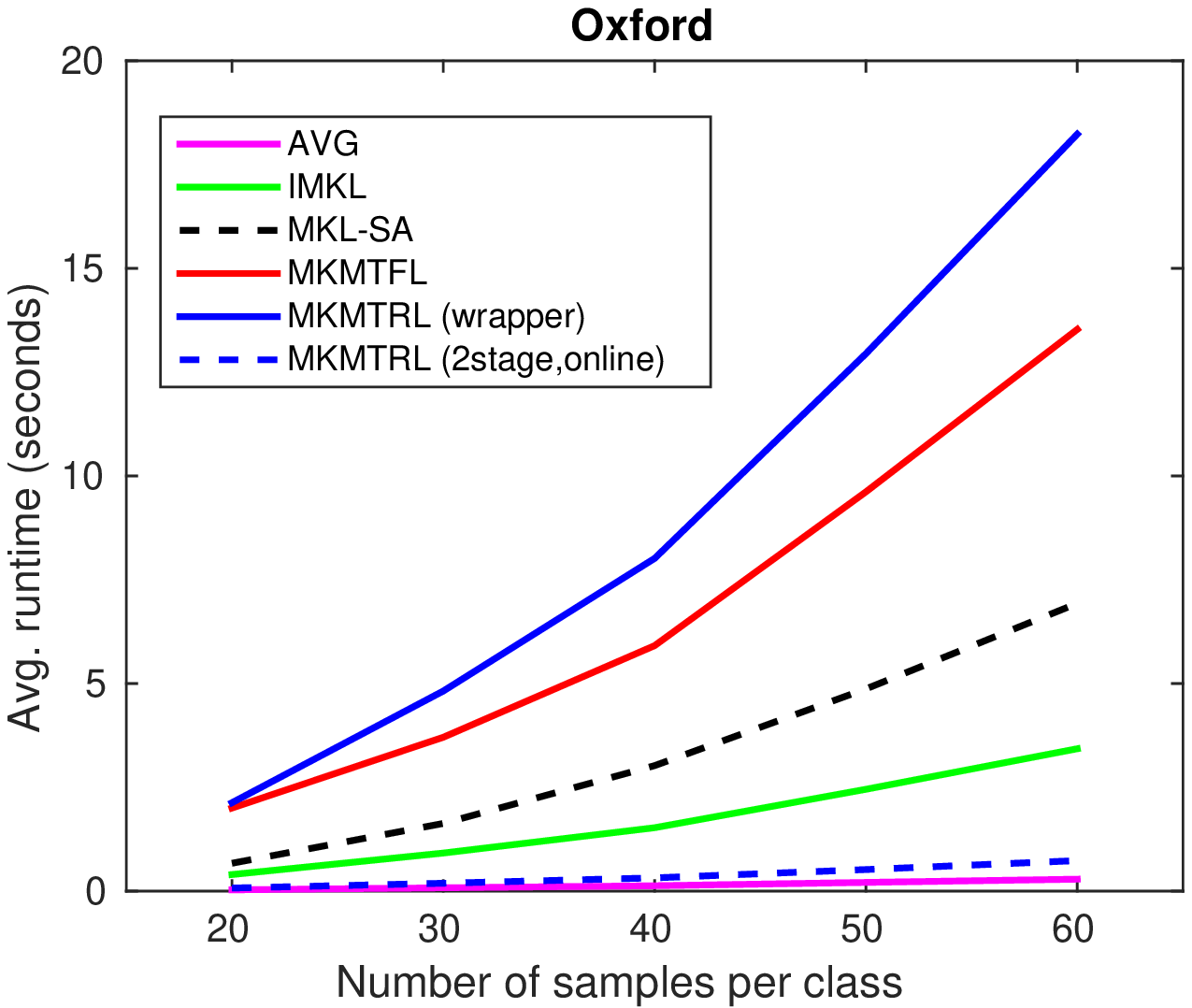}

	\caption{\textbf{Top}: Mean accuracy (left) and runtime (right) calculated for \textit{Caltech101} dataset with varying training set sizes. \textbf{Bottom}: Mean accuracy (left) and runtime (right) calculated for \textit{Oxford} dataset with varying training set sizes.}
	\label{fig:obj}
\end{figure*}

In addition to the baselines used before, we compare our algorithms with Multiple Kernel Learning by Stochastic Approximation (\textit{MKL-SA}) \cite{bucak2010multi}. \textit{MKL-SA} has a similar formulation to that of (\textit{MK-MTFL}), except that it sets $\lambda_{tk}=\lambda_t, \forall k$ in equation \ref{eq:mkmtfl}. At each time step, it samples one task, according to the multinomial distribution $Multi(\lambda_1, \lambda_2, \ldots, \lambda_T)$, to update it's model parameter, making it suitable for multitask learning with large number of tasks.

The results for \textit{Caltech101} and \textit{Oxford} are shown in Figure \ref{fig:obj}. The left plots show how the mean accuracy varies with respect to different training set sizes. The right plots show the average training time taken by each model with varying training set sizes. From the plots, we can see that \textit{MK-MTRL} outperforms all the other state-of-the art baselines in both \textit{Caltech101} and \textit{Oxford} datasets. But one may notice that the run-time of \textit{MK-MTFL} and \textit{MK-MTRL} grows steeply in the number of samples per class. Similar results are observed when we increase the number of tasks or number of base kernels per task.

Since both \textit{MK-MTFL} and \textit{MK-MTRL} require the base kernels in memory to learn the kernel weights and the task relationship matrix iteratively, this poses a serious computational burden and explains our need for efficient learning algorithm for multitask multiple kernel learning problems. We report \textit{MK-MTRL} with two-stage, online procedure as one of the baselines. On both \textit{Caltech101} and \textit{Oxford}, the two-stage procedure yields comparable performance to that of \textit{MK-MTRL}. 

The run-time complexity of two-stage, online \textit{MK-MTRL} learning is significantly better than almost all the baselines. Since \textit{AVG} takes the average of the task-specific base kernels, it has the lowest computational time. It is interesting to see that two-stage, online \textit{MK-MTRL} performs better than \textit{MKL-SA} both in terms of accuracy and running time. We believe that since \textit{MKL-SA} updates the kernel weights after learning a single model parameter, it takes more iterations to converge (in term of model parameters and the kernel weights).

\begin{table*}[ht]
	\begin{minipage}[b]{0.56\linewidth}
		\centering
		\renewcommand{\arraystretch}{1.5}
		\caption{Comparison for multiple kernel models using nMSE on SARCOS data}
		\label{tab:sarcos}
		\begin{tabular}{l|lll|}
			\cline{2-4}
			& STL                                                            & IMKL                                                            & MK-MTRL                                                                 \\ \hline
			\multicolumn{1}{|l|}{1st DOF} & \begin{tabular}[c]{@{}l@{}}0.0862 \\ $\pm$ 0.0033\end{tabular} & \begin{tabular}[c]{@{}l@{}}0.0838 \\ $\pm$ 0.0032\end{tabular} & \textbf{\begin{tabular}[c]{@{}l@{}}0.0717\\ $\pm$ 0.0075\end{tabular}}  \\
			\multicolumn{1}{|l|}{2nd DOF} & \begin{tabular}[c]{@{}l@{}}0.0996 \\ $\pm$ 0.0041\end{tabular} & \begin{tabular}[c]{@{}l@{}}0.0945\\ $\pm$ 0.0045\end{tabular}  & \textbf{\begin{tabular}[c]{@{}l@{}}0.0686\\ $\pm$ 0.0070\end{tabular}}  \\
			\multicolumn{1}{|l|}{3rd DOF} & \begin{tabular}[c]{@{}l@{}}0.0918 \\ $\pm$ 0.0042\end{tabular} & \begin{tabular}[c]{@{}l@{}}0.0871 \\ $\pm$ 0.0040\end{tabular} & \textbf{\begin{tabular}[c]{@{}l@{}}0.0649 \\ $\pm$ 0.0071\end{tabular}} \\
			\multicolumn{1}{|l|}{4th DOF} & \begin{tabular}[c]{@{}l@{}}0.0581 \\ $\pm$ 0.0021\end{tabular} & \begin{tabular}[c]{@{}l@{}}0.0514 \\ $\pm$ 0.0020\end{tabular} & \textbf{\begin{tabular}[c]{@{}l@{}}0.0298 \\ $\pm$ 0.0037\end{tabular}} \\
			\multicolumn{1}{|l|}{5th DOF} & \begin{tabular}[c]{@{}l@{}}0.1513 \\ $\pm$ 0.0063\end{tabular} & \begin{tabular}[c]{@{}l@{}}0.1405 \\ $\pm$ 0.0057\end{tabular} & \textbf{\begin{tabular}[c]{@{}l@{}}0.1070 \\ $\pm$ 0.0053\end{tabular}} \\
			\multicolumn{1}{|l|}{6th DOF} & \begin{tabular}[c]{@{}l@{}}0.2911 \\ $\pm$ 0.0094\end{tabular} & \begin{tabular}[c]{@{}l@{}}0.2822 \\ $\pm$ 0.0081\end{tabular} & \textbf{\begin{tabular}[c]{@{}l@{}}0.1835 \\ $\pm$ 0.0125\end{tabular}} \\
			\multicolumn{1}{|l|}{7th DOF} & \begin{tabular}[c]{@{}l@{}}0.0715 \\ $\pm$ 0.0025\end{tabular} & \begin{tabular}[c]{@{}l@{}}0.0628 \\ $\pm$ 0.0024\end{tabular} & \textbf{\begin{tabular}[c]{@{}l@{}}0.0457 \\ $\pm$ 0.0036\end{tabular}} \\ \hline
			\multicolumn{1}{|l|}{AVG}     & \begin{tabular}[c]{@{}l@{}}0.1214 \\ $\pm$ 0.0015\end{tabular} & \begin{tabular}[c]{@{}l@{}}0.1146 \\ $\pm$ 0.0013\end{tabular} & \textbf{\begin{tabular}[c]{@{}l@{}}0.0816 \\ $\pm$ 0.0028\end{tabular}} \\ \hline
		\end{tabular}
		\end{minipage}\hfill
        \begin{minipage}[b]{0.4\linewidth}
        \begin{minipage}{\linewidth}
        \centering
				\includegraphics[width=2.5in]{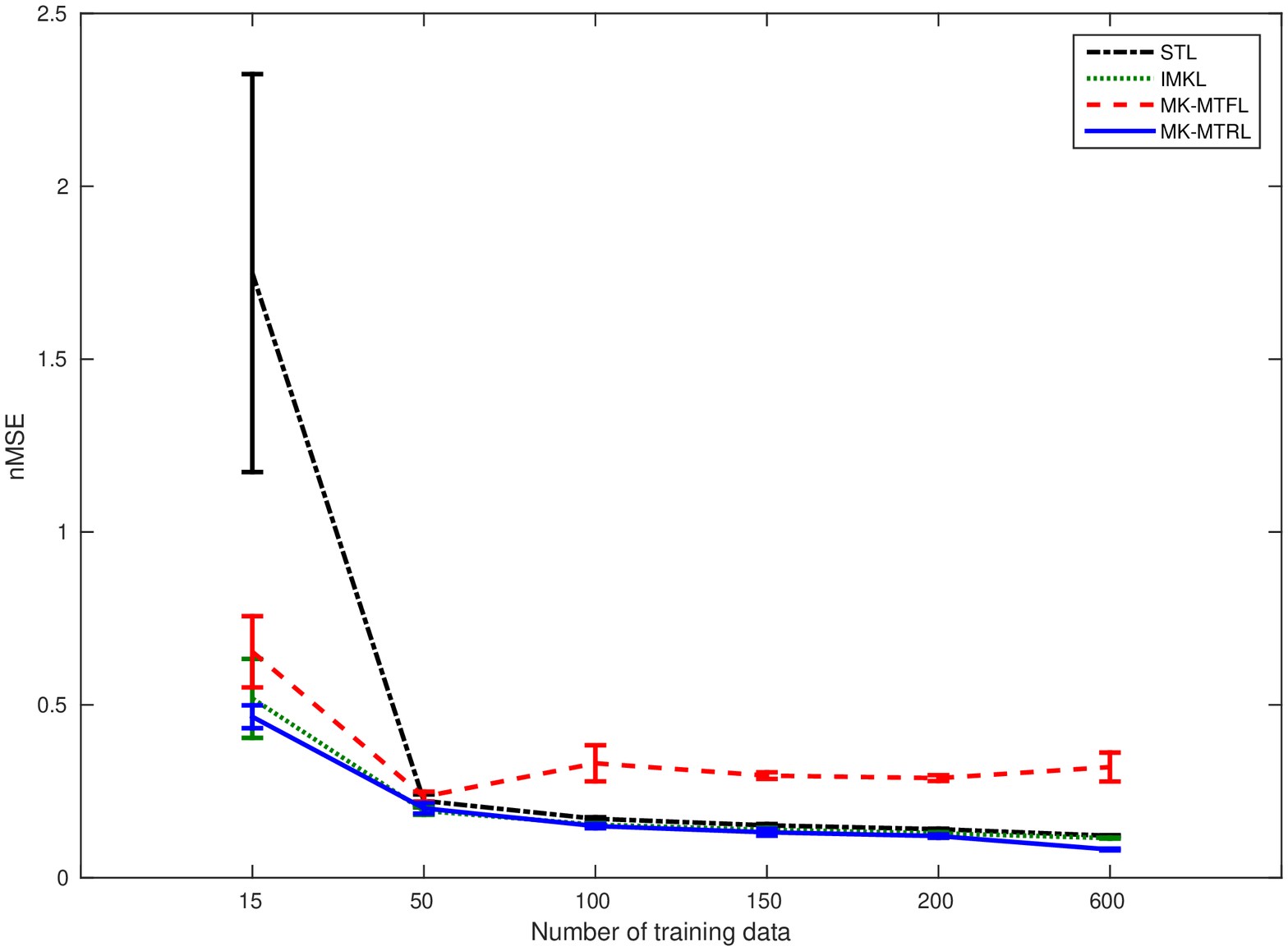}
				\captionof{figure}{nMSE vs Number of training example for SARCOS data}
				\label{fig:sarcos}

				\captionof{table}{Experiment on  \textit{school} dataset}
				\begin{tabular}{l|l|}
				\cline{2-2}
				& Explained Variance          \\ \hline
				\multicolumn{1}{|l|}{STL}    & 0.1883 $\pm$ 0.020          \\
				\multicolumn{1}{|l|}{IMKL}    & 0.1975 $\pm$ 0.017          \\
				\multicolumn{1}{|l|}{MK-MTFL} & 0.2024 $\pm$ 0.016          \\
				\multicolumn{1}{|l|}{MK-MTRL} & \textbf{0.2134 $\pm$ 0.016} \\ \hline
			\end{tabular}
        \end{minipage}
        \end{minipage}
	\end{table*}
	
	\subsection{Robot Inverse Dynamics}
We consider the problem of learning the inverse dynamics of a $7$-DOF \textit{SARCOS} anthropomorphic data \footnote{\url{http://www.gaussianprocess.org/gpml/data/}}. The dataset consists of 28 dimensions, of which first 21 dimensions are considered as features and the last 7 dimensions are used as outputs. We add an additional feature to account for the bias. There are 7 regression tasks and use  kernel ridge regression to learn the task parameters and kernel weights. The feature set includes seven joint positions, seven joint velocities and seven joint accelerations, which is used to predict seven joint torques for the seven degrees of freedom (DOF). We randomly sample $2000$ examples, of which $\{15,50,100,150,200,600\}$ are used for training sets and the rest of the examples are used for test set.

This dataset has been previous shown to include positive correlation, negative correlation and task unrelatedness and will be a challenging problem for baselines that doesn't learn the task correlation.

Following \cite{zhang2014regularization}, we use normalized Mean Squared error (nMSE), which is the mean squared error divided by the variance of the ground truth.  We generate $31$ base kernels from multivariate Gaussian kernels with $10$ varying bandwidth (based on the range of the data) and feature-wise linear kernel on each of the $21$ dimensions. We use linear kernel for single task learning. The results calculated for different training set size is reported in Figure \ref{fig:sarcos}. We can see that \textit{MK-MTRL} performs better than all the baselines. Contrary to the results report in \cite{jawanpuria2011multi}, \textit{MK-MTFL} performs the worst. As the model sees more data, it struggles to learn the task relationship and even performs worse than the single task learning.

	Moreover, we report the individual nMSE for each DOF in Table \ref{tab:sarcos}. It shows that \textit{MK-MTRL} consistently outperforms in all the tasks. Comparing the results to the one reported in  \cite{zhang2014regularization}, we can see that \textit{MT-MTRL} (with $0.0816$ AVG nMSE score) performs better than \textit{MTFL} and \textit{MTRL} (with $0.3149$ and $0.0912$ AVG nMSE scores respectively).

		\subsection{Exam Score Prediction}
		For completeness, we include the results for benchmark dataset in multi-task regression \footnote{\url{http://ttic.uchicago.edu/~argyriou/code/mtl_feat/school_splits.tar}}. The \textit{school} dataset consists of examination scores of $15362$ students from $139$ schools in London. Each school is considered as a task and the feature set includes  the year of the examination, four school-specific and three student-specific attribute. We replace each categorical attribute with one binary variable for each possible attribute value, as in \cite{argyriou2008convex}. This results in $26$ attributes with additional attribute to account for the bias term. We generate univariate Gaussian kernel with 13 varying bandwidths from each of the $26$ attributes as our base kernels. Training and test set are obtained by dividing examples of each task into $60$\%-$40$\%. We use  explained variance as in \cite{argyriou2008convex}, which is defined as one minus nMSE. We can see that \textit{MK-MTRL} is better than both \textit{IMKL} and \textit{MK-MTFL}.